# A Measure of Decision Flexibility

Ross D. Shachter
Department of
Engineering-Economic Systems
Terman Engineering Center
Stanford University
Stanford, CA 94305-4025
shachter@camis.stanford.edu

Marvin Mandelbaum
Dept. of Computer Science and Math
Atkinson College, York University
4700 Keele Street
North York, Ontario M3J 1P3
CANADA
mandel@sol.yorku.ca

## Abstract

We propose a decision-analytical approach to comparing the flexibility of decision situations from the perspective of a decision-maker who exhibits constant risk-aversion over a monetary value model. Our approach is simple yet seems to be consistent with a variety of flexibility concepts, including robust and adaptive alternatives. We try to compensate within the model for uncertainty that was not anticipated or not modeled. This approach not only allows one to compare the flexibility of plans, but also guides the search for new, more flexible alternatives.

## 1 INTRODUCTION

In any systematic approach to decision-making, whether it be decision analysis, AI planning, or corporate strategy, a desirable feature in any plan is "flexibility." Unfortunately, as tempting as the concept of flexibility is, it has been hard to define. In this paper we propose a simple, decision-analytic approach to comparing the flexibility of two plans. This approach is consistent with the variety of concepts in the literature.

An early classic work on flexibility is Stigler(1939). This paper compares two potential configurations for the operation of a factory, with two corresponding cost curves, as shown in Figure 1. One curve, $C1$, can acheive a lower cost, but it is quite sensitive to the uncertain quantity. The other curve, $C2$, is less sensitive to the uncertainty, but might cost more than $C1$. This is an example of what we call *robust flexibility*, since our concern is developing a plan that will function well in the face of uncertainty. To analyze Stigler's problem, it is not enough to know the cost curves; we need a probability distribution over the uncertain quantity, too.

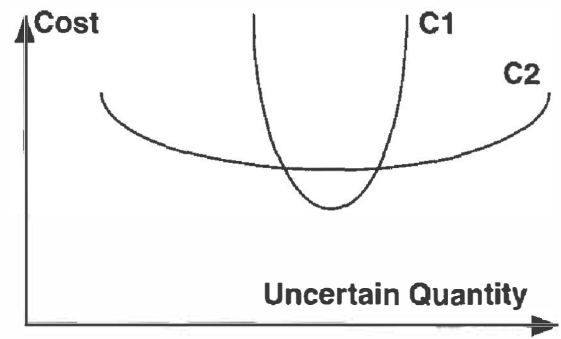

Figure 1: A production facility can operate at two cost curves as a function of an uncertain quantity. Although curve $C1$ can acheive a lower cost, $C2$ is less sensitive to the uncertainty. This is an example of robust flexibility.

Some more sophisticated approaches to flexibility involve the subtle interaction of information and decisions (Chávez and Shachter 1995; Jones and Ostroy 1984; Mandelbaum 1978; Marschak and Miyasawa 1968; Marschak and Nelson 1962). These frameworks involve two sequential decisions, a *commitment* and a *reaction*, with an intervening obseration, as represented by the influence diagram shown in Figure 2. Usually some commitments offer more or better reaction alternatives, but at a cost. This is what we call *adaptive flexibility*, since the flexibility comes from the opportunity to react in light of the observation.

Another school of thought is that decision analysis automatically and implicitly incorporates flexibility (Merkofer 1977). This assumes that we have modeled all of the choices and information available. This approach would subsume both robust and adaptive flexibility, but seems to miss something critical that motivates modelers to consider issues of flexibility in the first place. That is, the decision analytic approach assumes that our model is complete.

In this paper, we follow the decision analysis tradition, but recognize that there is some uncertainty that we were unable to model, perhaps because we could not anticipate it. Reasoning by analogy from the uncer-tainty we are able to model, we guess how much uncertainty is "missing" from our model, and we "add" it directly to our value model. This allows us to compare different plans or decision situations, regardless whether they fit the templates for robust or adaptive flexibility. Although the resulting measure is approximate, it suggests new alternatives we can introduce to increase flexibility.

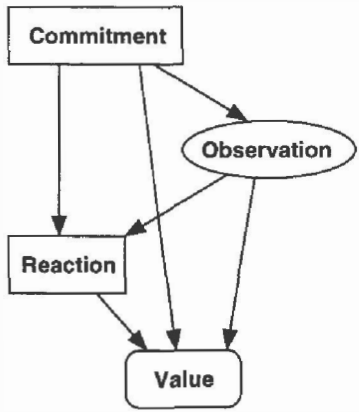

Figure 2: An adaptive plan is represented in this influence diagram. After a commitment decision is made, an observation is revealed, and then a reaction decision is made. The observation can depend on the commitment, and the value of the situation can depend on the decisions and the observation.

To illustrate these concepts, consider the management of an investment portfolio. An inflexible investment might be one in which the capital is locked in place for a long term. A robust investment might be in an instrument that performs well in a variety of market conditions. On the other hand, an adaptive investment might be one that can be shifted easily in response to changing markets. Decision analysis provides the machinery for us to explicitly balance the value of this flexibility against any premium in cost. Finally, we might chose to magnify the uncertainty in our decision analysis model, if we believe that there are important uncertainties we have failed to model.

In Section 2 we will explore this concept in more detail. Section 3 develops notation and technical results. We combine those in Section 4 to obtain our definition of flexibility orders and investigate its implications. Finally, we discuss our conclusions and some directions for further research.

## 2 CONCEPTS

In this section we describe the concepts behind our approach to flexibility. Coupled with the technical results in the next section, this leads to the flexibility orders in Section 4.

We can identify several criteria that we would like our approach to satisfy. First, we would like it to operate on a large variety of general decision models. Ideally, it should be applicable to models represented by the decision analysis type of decision trees (Raiffa 1968) or influence diagrams (Howard and Matheson 1984). Second, it should be conceptually simple. Third, it should be useful for suggesting or stimulating the generation of new plans or alternatives. Fourth, it should be general enough to incorporate most of the concepts currently in the literature. Fifth, it should agree with our common sense; for example, adding a new alternative to a plan should never decrease its flexibility.

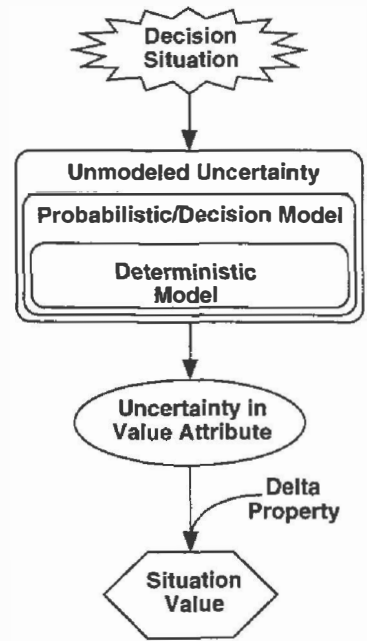

Figure 3: The valuation of a decision situation is illustrated by a flow chart. Models of the situation are shown by the rounded rectangles, necessarily limited in accuracy. The distribution over the value attribute can then be computed. When the decision-maker satisfies the delta property, it is then easy to compute an equivalent certain attribute value for the decision situation.

We believe that there is an underlying theme behind the concepts of flexibility in the literature, namely, that a plan must be able to perform well under unanticipated or unmodeled uncertainty. Consider the flow chart shown in Figure 3. Under any decision situation, we can model the performance of a system with varying degrees of accuracy. The simplest model might be deterministic; we can acheive more accuracy by considering uncertainty. Regardless, there will be considerable uncertainty that remains unmodeled, either because it was too complex, or because we could not anticipate it.

One challenge is how to recognize and estimate this unmodeled uncertainty. It is difficult to incorporate it into the model, but since all we really care about is the uncertainty in the value attribute, we don't need to add the uncertainty to the model. Instead, we directly manipulate the value attribute distribution to incorporate the uncertainty we believe is missing from the model. Finally, if we have additional structure in the value model, as assumed in the next section, we can simplify the analysis of the value of the decision situation.

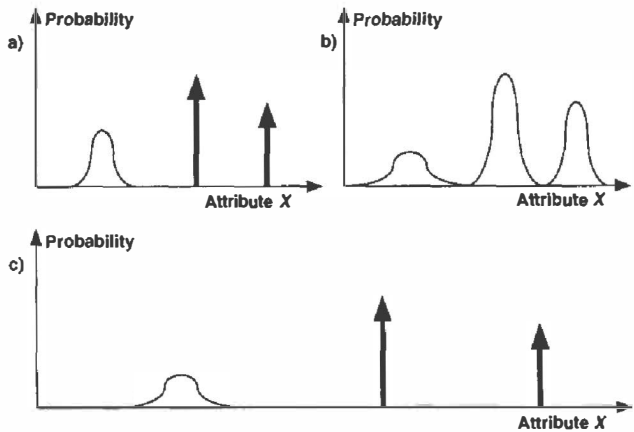

Figure 4: Probability distributions over the model attribute $X$ before and after manipulations.

There are a couple of ways we can introduce uncertainty into the value attribute distribution. Consider the probability distributions over value attributes shown in Figure 4. The modeled distribution in shown in Figure 4a. A natural way to introduce uncertainty into the value distribution is to add independent uncertainty to obtain the distribution shown in Figure 4b. Note that the probability masses become densities, and the the density becomes more diffuse. Another transformation is to rescale the attribute axis to obtain the distribution shown in Figure 4c. In this case, probability masses stay masses but densities become more diffuse.

This approach is based on several key assumptions. First, we can make up for the unmodeled uncertainty by adding some to the model. Second, we can use the uncertainty in the model as a guide for where to add the additonal uncertainty. Third, the model is robust enough to provide sensible outputs when it is distorted beyond its designed range. These assumptions are hard to justify, but reasonable if we are to exercise the model to handle the unforeseen. Keeping them in mind, we can recognize that perhaps the most valuable use of the flexibility measure might be in inspiring the generation of new alternatives, that is, suggesting superior plans.

## 3 NOTATION AND ANALYTICAL RESULTS

In this section, we introduce fundamental concepts of decision-making and the characterization of the value of an uncertain decision situation. We derive a new, simple, and useful result that will allow us to recognize flexibility orders for particular types of decision-makers.

The decision-maker takes actions that affect her world in uncertain ways, and she chooses actions by comparing their anticipated effects. Bayesian decision analysis provides a coherent rational framework for analyzing her choices. This is not a _descriptive_ technique predicting the behavior of the decision-maker in practice, but rather a _normative_ approach consistent with a set of conditions that she might reasonably choose to satisfy.

A _decision_ is an irrevocable allocation of resources, usually framed as the selection of an element from a set of possible choices called _alternatives_. To be a decision, the choice has to be available to the decision-maker, and it has to be a real commitment. If she can change her mind at no cost she has not yet made a decision. Conversely, if the choice has not yet been made officially, but there is an informal understanding that would make it costly to change, then a decision has been made already. This corresponds to the flexibility with respect to observations in Merkofer(1977).

After making her choice, the decision-maker faces an uncertain future. She can analyze any possible situation by considering each possible state and assigning it a _von Neumann-Morgenstern utility_ $u$, such that if two situations have uncertain utilities $U_1$ and $U_2$, respectively, she would prefer the former if and only if $E\{U_1\} > E\{U_2\}$.

In this paper, we assume that she can also characterize her future in terms of a single variable or _attribute_ of the problem. Her beliefs about the uncertain attribute $X$ we will call her _prospect_. The most common single attribute is money, although other attributes might be more appropriate for particular problems, such as the probability of successful mission completion for an intelligent agent. Nonetheless, for simplicity the rest of the paper assumes that the single attribute is money.

If the decision-maker is always willing to place a monetary value on any prospect, she is said to have a _monetary equivalence_ for prospects. If she has a monetary equivalence, there is some function $\$(u)$ called the _willingness to pay_, defined as the most she would pay to have a prospect with utility $u$. Assume that $\$(u)$ is strictly increasing, so it has a well defined inverse function, $u_\$(\$)$, the _utility for money_.

A decision-maker with monetary equivalence is said to satisfy the _delta property_ if, for any uncertain monetary prospect $X$ and constant payment $c$,

$$\$(E\{u_\$(X+c)\}) = \$(E\{u_\$(X)\}) + c.$$

In that case, she is willing to pay $\$(u_2) - \$(u_1)$ money to go from a prospect with utility $u_1$ to a prospect with utility $u_2$. The delta property greatly simplifies calculations of monetary equivalence, but it also im-plies some strong restrictions on $\$(u)$ and $u_\$(\$)$. The following theorem is well known (Pratt 1964) and fundamental to our analysis.

**Theorem 1 (Delta Property)** *Given a monetary equivalent decision-maker with a utility function for money $u_\$(x)$ that is twice continuously differentiable and $u'_\$(x) > 0$, she satisfies the delta property if and only if her absolute risk-aversion is a constant $r$ and*

$$u_\$(x) = \begin{cases} x & \text{if } r = 0 \\ \frac{1 - exp(-rx)}{1 - exp(-r)} & \text{if } r \neq 0 \end{cases}$$

*assuming, without loss of generality, that $u_\$(0) = 0$ and $u_\$(1) = 1$. The more general class of utility functions, given by $a + bu_\$(x)$ where $b > 0$, reveals the same preferences among prospects for all choices of $a$ and $b$.*

If she satisfies the delta property, the decision-maker's attitude toward risk is characterized by her constant *absolute risk-aversion*, $r = -u''_\$(x)/u'_\$(x)$. Her willingness to pay for any uncertain monetary prospect $X$ is called her *certain equivalence* for $X$, $CE(X|r)$, and it is defined as

$$CE(X|r) = \$(E\{u_\$(X)\}).$$

In general, it is different from the expected value of the prospect, $E\{X\}$, and they are related by the approximation,

$$\$(E\{u_\$(X)\}) \approx E\{X\} + \frac{r\text{Var}\{X\}}{2},$$

which is exact when $X$ has a Gaussian (or normal) distribution. If she is indifferent between any prospect and its expectation then $r = 0$ and she is said to be *risk-neutral*. If she never prefers any uncertain monetary prospect to its expectation then $r > 0$ and she is said to be *risk-averse*. In general, concave utility functions are risk-averse and linear utility functions are risk-neutral. Using Theorem 1, we can simplify the certain equivalence formula as follows.

**Theorem 2 (Certain Equivalence)** *Given a monetary equivalent decision-maker satisfying the delta property with risk-aversion $r > 0$ and any uncertain monetary prospect $X$,*

$$CE(X|r) = -\frac{1}{r} \ln E\{e^{-rX}\}.$$

**Proof:**

$$\begin{aligned}
CE(X|r) &= \$(E\{u_\$(x)\}) \\
&= -\frac{1}{r} \ln(1 - (1 - e^{-r})E\{\frac{1 - e^{-rX}}{1 - e^{-r}}\}) \\
&= -\frac{1}{r} \ln(1 - E\{1 - e^{-rX}\}) \\
&= -\frac{1}{r} \ln(E\{e^{-rX}\}) \quad \square
\end{aligned}$$

For the comparison of flexibility among decision situations, we would like to characterize the decision-maker's willingness to pay for prospects after a linear transformation. This can be done simply when the decision-maker satisfies the delta property.

**Theorem 3 (Linear Transformation)**
*Given a monetary equivalent decision-maker satisfying the delta property with risk-aversion $r$, independent uncertain monetary prospects $X$ and $Z$, and positive constant $k$,*

$$CE(kX + Z|r) = kCE(X|kr) + CE(Z|r).$$

**Proof:**

$$\begin{aligned}
CE(kX + Z|r) &= -\frac{1}{r} \ln(E\{e^{-r(kX+Z)}\}) \\
&= -\frac{1}{r} \ln(E\{e^{-rkX}e^{-rZ}\}) \\
&= -\frac{1}{r} \ln(E\{e^{-rkX}\}E\{e^{-rZ}\}) \\
&= -\frac{\ln(E\{e^{-rkX}\})}{r} - \frac{\ln(E\{e^{-rkX}\})}{r} \\
&= kCE(X|kr) + CE(Z|r) \quad \square
\end{aligned}$$

As a result of Theorem 3, we have a simple rule for comparing linearly transformed prospects when the decision-maker satisfies the delta property.

**Corollary 1 (Transformation Comparison)**
*Given a monetary equivalent decision-maker satisfying the delta property with risk-aversion $r$, uncertain monetary prospects $X$ and $Y$, uncertain monetary prospect $Z$ independent of $X$ and $Y$, and positive constant $k$, she prefers prospect $kX + Z$ to $kY + Z$ if and only if $CE(X|kr) > CE(Y|kr)$.*

## 4 FLEXIBILITY ORDERS

We are now ready to state precisely what it means for one decision situation to be more flexible than another. In this section, we assume that the decision-maker is monetary equivalent satisfying the delta property with risk-aversion $r > 0$. In that case, we can apply the results of the previous section to the concepts introduced earlier to obtain a rule for comparing situations. We explore the properties of this rule and show that it satisfies many of the desiderata we identified earlier.

Our approach to comparing decision situations is to attenuate the uncertainty in the prospects $X$ and $Y$ by varying a parameter $k$. As $k$ increases from 1, the uncertainty in the value distribution is magnified. Two kinds of uncertainty can be added to the distribution, an independent uncertainty and a rescaling of the attribute scale, as shown in Figure 4. Given an uncertain monetary prospect $Z$ which is independent of $X$ and $Y$ (but might depend on $k$), we transform $X$ to $kX + Z$ to compare flexibilities. The decision-maker considers $X$ more flexible than $Y$ if she prefers $kX + Z$ to

$kY + Z$ for all $k \geq 1$. If she satisfies the delta property, the effects of these changes are easy to analyze, using the results in Corollary 1. Adding uncertainty seems natural, but doesn't change her relative ordering over situations. Rescaling the attribute does change her ordering, by effectively rescaling her risk-aversion. These results are incorporated into the following definitions.

Given two uncertain monetary prospects $X$ and $Y$, a monetary equivalent decision-maker satisfying the delta property with risk-aversion $r > 0$ is said to find $X$ <u>(strictly) more flexible</u> than $Y$ if there is some positive $K > 0$ such that

$$\mathrm{CE}(X|kr) \geq (>)\mathrm{CE}(Y|kr) \text{ for all } k \geq K.$$

She is said to find that $X$ <u>(strictly) dominates $Y$ in flexibility</u> if she finds that $X$ is (strictly) more flexible than $Y$ with $K = 1$, that is,

$$\mathrm{CE}(X|kr) \geq (>)\mathrm{CE}(Y|kr) \text{ for all } k \geq 1.$$

By this definition, if $X$ (strictly) dominates $Y$ in flexibility then $X$ is (strictly) more flexible than $Y$.

Consider the graphs shown in Figure 5. In Figure 5a and Figure 5b, the uncertain monetary prospect $Y$ is strictly more flexible than the uncertain monetary prospect $X$, while in Figure 5d, $X$ is more flexible than $Y$, but not strictly. There is dominance shown in Figure 5a, where $Y$ strictly dominates $X$ in flexibility, and Figure 5d, where $X$ dominates $Y$ in flexibility. There is no dominance shown in Figure 5b. It is quite possible to have curves like those shown in Figure 5c, where neither $X$ nor $Y$ is more flexible, and thus neiher can dominate.

Most of the properties of the flexibility comparison follow from properties of the certain equivalent, $\mathrm{CE}(X|kr)$. The certain equivalent is in units of an attribute, which we have assumed is money. As a function of $k$, $\mathrm{CE}(X|kr)$ is continuous and nondecreasing (assuming $r > 0$).

If $X$ is deterministic, that is, it takes on a value for certain, then $\mathrm{CE}(X|kr) = \mathrm{CE}(X|r)$ for all $k > 0$. The worst case for $X$, the smallest possible value of $X$, is approached by $\mathrm{CE}(X|kr)$ as $k$ increases. More formally,

$$\lim_{k \to \infty} \mathrm{CE}(X|kr) \leq x \text{ if } \Pr\{X \leq x\} > 0.$$

Therefore, if $X$ and $Y$ are two monetary prospects between which the decision-maker is indifferent and $Y$ is deterministic, then $Y$ dominates $X$ in flexibility.

We can now return to the desiderata from Section 2 to see (surprise! surprise!) that they are satisfied by the proposed approach. It is conceptually simple and can be applied to a wide variety of decision models, including decision trees and influence diagrams. In fact, it can even be applied to compare the flexibility at different nodes in the same decision tree. Since the certain equivalent never decreases when an alternative is added, it satisfies our intuition in that flexibility does not decrease in that case. Of course, it allows the decision-maker to test new alternatives easily and might help her conceive of them, too.

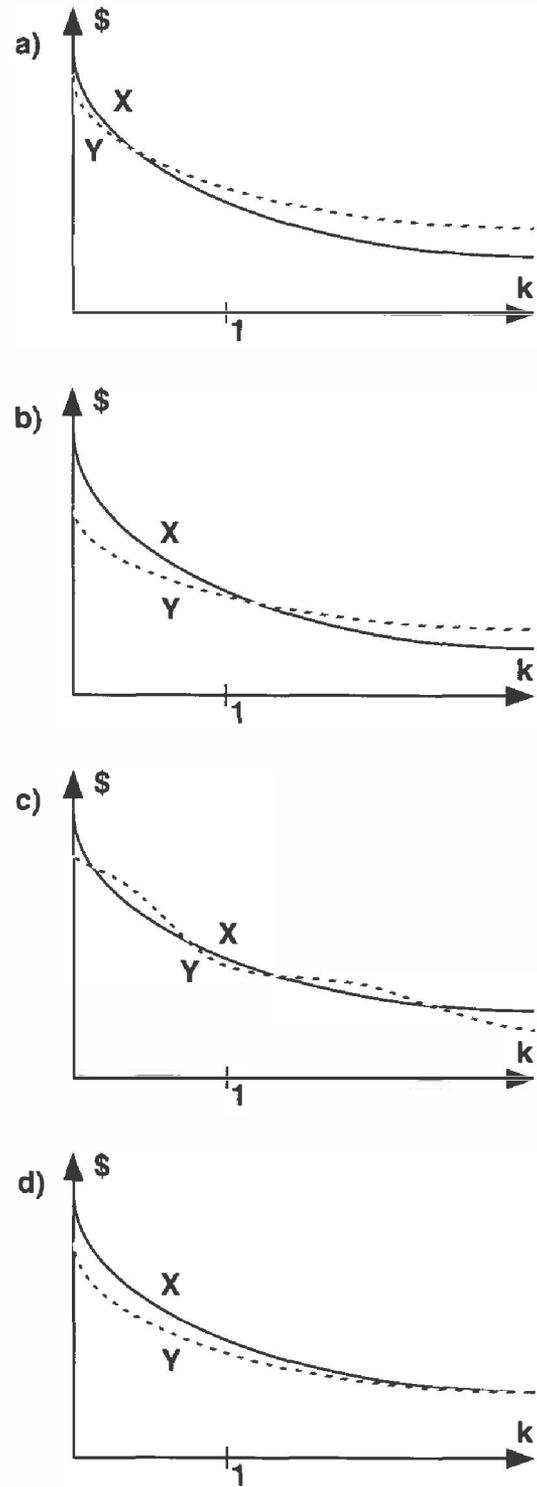

Figure 5: Example graphs of certain equivalents as a function of $k$ for two uncertain monetary prospects, $X$ and $Y$. The certain equivalent for $X$ is drawn with a solid line while the certain equivalent for $Y$ is drawn with a dashed line. The graphs illustrate, respectively, $Y$ strictly dominating $X$ in flexibility, $Y$ strictly more flexible than $X$, neither more flexible, and $X$ dominating $Y$ in flexibility.

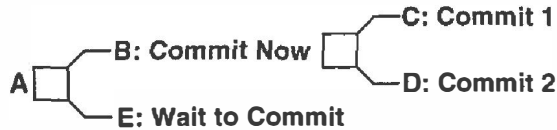

Figure 6: This is a decision analysis type decision tree in which every node represents a decision. We can choose to commit to either 1 ($C$) or 2 ($D$), or choose between them now ($B$) or choose between them later ($E$). The situation where we can decide whether to choose now or later is $A$.

What remains to be shown is that this approach is general enough to incorporate most of the concepts of flexibility in the literature. This follows from the decision analytical nature of the approach. Consider the notion of robust flexibility introduced by Stigler(1939) and illustrated in Figure 1. As $k$ is increased we would expect that the certain equivalent for the situation with cost curve $C1$ will decrease more rapidly than the situation with cost curve $C2$. Thus, we obtain the intuitive result that $C2$ is strictly more flexible than $C1$, although not necessarily dominant.

Consider the adaptive flexibility case represented by the influence diagram shown in Figure 2. For each possible decision strategy there is a flexibility curve, and we might expect some curves to be more flexible than others. Often these will be the most adaptive plans, but it is quite possible that a robust plan could be the most flexible. Thus we can represent adaptive flexibility but we are not restricted to it.

Another example of flexibility comparison corresponds to the sequential decision tree shown in Figure 6. Decision $B$ lets us choose between decision $C$ or decision $D$, so the certain equivalent at $B$ is at least as big as the certain equivalent at either $C$ or $D$ for all $k$. Therefore, $B$ dominates both $C$ and $D$ in flexibility. By the same logic, $A$ dominates $B$, $C$, $D$, and $E$ in flexibility. If there is no cost to waiting to commit, then $E$ dominates $B$, $C$, and $D$ in flexibility.

The analysis is further simplified if the uncertain monetary prospect $X$ is distributed with a Gaussian distribution, characterized by its mean, $\mathrm{E}\{X\}$, and variance, $\mathrm{Var}\{X\}$. In this case, the certain equivalent $\mathrm{CE}(X|kr)$ is linear in $k$,

$$\mathrm{CE}(X|kr) = \mathrm{E}\{X\} - \frac{\mathrm{Var}\{X\}}{2}kr.$$

Note that the certain equivalent is unbounded as $k$ increases unless the prospect is deterministic. That is because there is probability mass at arbitrarily low values, so the worst case certain equivalent is unbounded.

If, in the Stigler example, the monetary prospects $C1$ and $C2$ have Gaussian distributions, then they might be represented by the graphs in either Figure 7a or Figure 7b. In both cases, $C2$ is strictly more flexible than $C1$ and in Figure 7b $C2$ strictly dominates $C1$ in flexibility. Similarly, if every monetary prospect in the adaptive example has a Gaussian distribution, then it might be represented by the graph in Figure 7c. If one curve has a flatter slope than another then it is more flexible. There can be a number of curves that are optimal for some $k$, but some of the curves might not be optimal for any $k$.

## 5 CONCLUSIONS AND FUTURE RESEARCH

In this paper, we have proposed a simple decision-analytic approach to comparing the flexibility of different plans or decision situations. This is particularly useful for automated reasoning tasks where we have limited resources for model construction and analysis (Horvitz 1990). Under such circumstances, we need a robust estimate of the uncertainty which has gone unmodeled, precisely what our measure of flexibility provides.

Our approach behaves according to our intuitive notions of flexibility as well as the concepts in the literature, without imposing severe restrictions on the types of models. Because the method forces a model beyond its design parameters, it can introduce additional modeling inaccuracy. Therefore, the "stretched" model should not be examined too precisely, but rather used for the critical creative task of stimulating and generating new, more flexible alternatives.

A natural direction for future research would relax the assumption that the decision-maker satisfies the delta property. To increase the uncertainty in the value attribute distribution as a function of parameter $k$, we could transform the uncertain prospect from $X$ to $kX + Z$ where $Z$ is an uncertain prospect which is independent of $X$, although it could depend on $k$. It is just not clear to us at this point what would be an appropriate value for $Z$. If a deterministic value were chosen for $Z$, then this approach would be easy to perform for general decision problems.

The proposed framework looks at different plans as we increase $k$ from 1. Some further research might investigate the value of considering values of $k$ less than 1. There might even be some insights to be gained by considering negative values of $k$, turning the decision-maker into a risk-seeker.

In this paper we have assumed that the decision-maker is risk-averse, but we could imagine modeling the behavior of a risk-neutral decision-maker, that is a monetary equivalent decision-maker satifying the delta property with absolute risk-aversion $r = 0$. This poses a couple of problems for our approach. First, there would be no sense in applying the method without any modifications, since $\mathrm{CE}(X|0k) = \mathrm{CE}(X|0)$. Second, in building a model for a risk-neutral decision-maker, there is no need to introduce much of the uncertainty, since it is not relevant for making the decisions. Thus such a model would be quite brittle with respect to changes in $r$.

## Acknowledgements

We benefited greatly from the comments of the anonymous referees and many colleagues, most notably James Matheson, John Mark Agosta, and Thomas Chávez.

## References

Chávez, T. and R. D. Shachter. "Decision Flexibility." In **Uncertainty in Artificial Intelligence: Proceedings of the Eleventh Conference**, eds. P Besnard and S Hanks. 77-86. San Mateo, CA: Morgan Kaufmann, 1995.

Horvitz, E. "Computation and Action Under Bounded Resources." PhD thesis, Stanford University, 1990.

Howard, R. A. and J. E. Matheson. "Influence Diagrams." In **The Principles and Applications of Decision Analysis**, eds. R. A. Howard and J. E. Matheson. II. Menlo Park, CA: Strategic Decisions Group, 1984.

Jones, R. A. and J. Ostroy. "Flexibility and Uncertainty." **Review of Economic Studies** L1 (1984): 13-32.

Mandelbaum, M. "Flexibility in Decision Making: An Exploration and Unification." PhD, Department of Industrial Engineering, University of Toronto, 1978.

Marschak, J. and K. Miyasawa. "Economic Comparability of Information Systems." **International Economic Review** 9 (2 1968): 137-173.

Marschak, T. and R. Nelson. "Flexibility, Uncertainty, and Economic Theory." **Macroeconomic** 14 (1962): 42-58.

Merkofer, M. W. "The Value of Information Given Decision Flexibility." **Management Science** 23 (7 1977): 716-727.

Pratt, J. W. "Risk Aversion in the Small and in the Large." **Econometrica** 41 (1964): 35-39.

Raiffa, H. **Decision Analysis**. Reading, MA: Addison-Wesley, 1968.

Stigler, G. "Production and Distribution in the Short Run." **Journal of Political Economy** 47 (1939): 305-329.

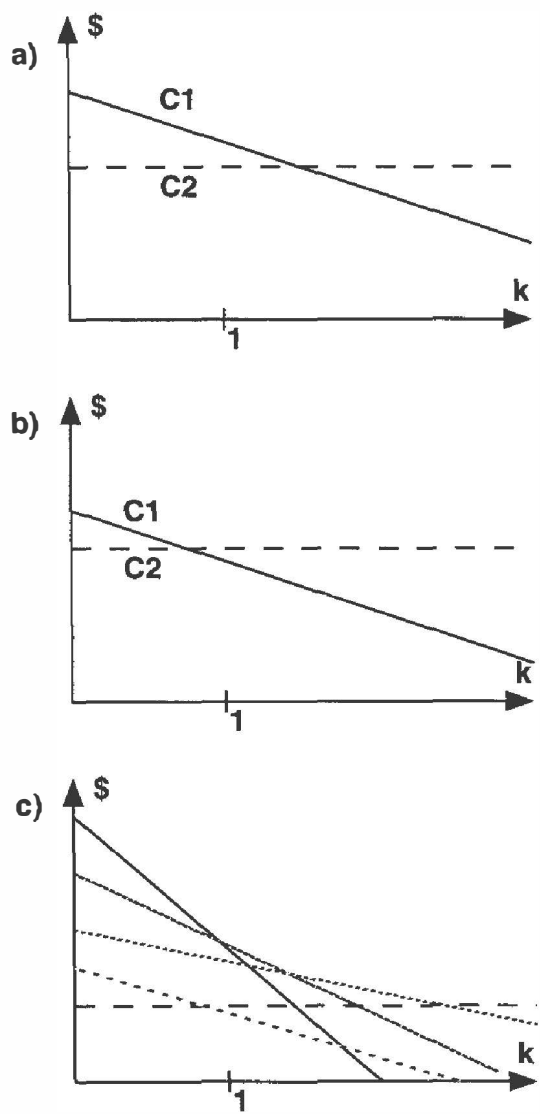

Figure 7: When the monetary prospects have Gaussian distributions, they are represented by lines in the graph. Figure 7a and Figure 7b represent cases of the Stigler example while Figure 7c represents an instance of the adapative example.